\ifdefined\directlua
  \DeclareFontFamily{OT1}{ptm}{}
  \DeclareFontShape{OT1}{ptm}{m}{n}{<->cmr10}{}
  \DeclareFontShape{OT1}{ptm}{m}{it}{<->cmti10}{}
  \DeclareFontShape{OT1}{ptm}{m}{sl}{<->cmsl10}{}
  \DeclareFontShape{OT1}{ptm}{b}{n}{<->cmbx10}{}
  \DeclareFontShape{OT1}{ptm}{bx}{n}{<->cmbx10}{}
  \DeclareFontShape{OT1}{ptm}{b}{it}{<->cmbxti10}{}
  \DeclareFontShape{OT1}{ptm}{bx}{it}{<->cmbxti10}{}
\fi
\documentclass[conference,letterpaper,10pt]{IEEEtran}

\usepackage{array}
\usepackage{amsmath}
\usepackage{amsfonts}
\usepackage{booktabs}
\usepackage{cite}
\usepackage{graphicx}
\usepackage[hidelinks]{hyperref}
\usepackage{tabularx}
\ifdefined\directlua
  \usepackage{fontspec}
\fi

\providecommand{\degree}{\ensuremath{^{\circ}}}
\hypersetup{hypertexnames=false}

\begin{document}

\title{Risk-Adaptive Edge--Cloud Visual Reasoning for Communication-Efficient Autonomous Driving}


\author{
\IEEEauthorblockN{
Meng Ma\IEEEauthorrefmark{1},
Shuyang Li\IEEEauthorrefmark{1},
Naigang Wang\IEEEauthorrefmark{2},
and Ruimin Ke\IEEEauthorrefmark{1}\textsuperscript{*}
}
\IEEEauthorblockA{
\IEEEauthorrefmark{1}Rensselaer Polytechnic Institute, Troy, NY, USA\\
\{mam7, lis36, ker\}@rpi.edu
}
\IEEEauthorblockA{
\IEEEauthorrefmark{2}IBM T. J. Watson Research Center, USA\\
nwang@us.ibm.com
}
\IEEEauthorblockA{
\textsuperscript{*}Corresponding author
}
}

\maketitle

\begin{abstract}
Cloud-hosted vision--language models (VLMs) offer greater contextual reasoning capabilities than smaller onboard models, but frequent visual uploads increase communication overhead and add network and inference latency to tactical decisions. We present a risk-adaptive edge--cloud architecture in which onboard traffic assessment determines when cloud reasoning is requested. An onboard VLM and a lightweight detector capture temporal traffic conditions and path-relative hazards for conservative local response and selective cloud access. The cloud model provides tactical advice, while validation, vehicle control, and automatic emergency braking remain local. In CARLA experiments, our method matched the task success rate of periodic cloud access while reducing cloud requests by 54.1\% and recording fewer automatic emergency braking (AEB) activations. In a delayed-roadwork ablation, semantic events triggered requests before the next scheduled audit. Across three emulated network profiles, the method continued to reduce cloud traffic, although lane changes took longer than with periodic access. Onboard traffic assessment therefore served as a practical trigger for selective VLM inference in these experiments.
\end{abstract}

\section{Introduction}

Autonomous driving requires timely local action and contextual reasoning under
finite onboard resources. Onboard computation avoids network dependence but
constrains model capacity and energy use, whereas remote execution can access
larger models at the cost of image transmission, latency, and network
variability \cite{gong2023edge,krentsel2024managing}. Edge intelligence and V2X
systems therefore motivate keeping time-critical functions near the vehicle
and invoking remote resources selectively
\cite{huang2023v2x,kang2017neurosurgeon,xu2023survey,zhu2023cloud,duan2021joint}.

This tradeoff is particularly important for vision--language models (VLMs).
Cloud-scale VLMs can interpret traffic scenes, reason about intent, and express
driving decisions
\cite{cui2024large,xu2024drivegpt4,mao2023gpt,cui2024personalized}, but dense
remote inference repeatedly transmits visual data and can return advice after
the scene has changed. Compact VLMs are more suitable for onboard inference,
yet their reduced capacity can limit open-ended and temporal reasoning
\cite{sharshar2025vision,cai2024self,chen2024omnivlm,liu2024lightweight};
driving evaluations report related spatio-temporal limitations
\cite{fruhwirth2025stsbench}. Model placement therefore depends not only on
model scale, but also on whether a traffic event warrants the communication
and response-time cost of cloud reasoning.

Selective remote inference also requires a clear authority model.
Split-computing and cloud-assisted systems partition computation or upload
selected observations
\cite{matsubara2022split,teerapittayanon2016branchynet,chen2024edge}, but a
driving stack must additionally specify which decisions remain local, how
delayed advice is validated, and when emergency logic takes precedence. We
study whether structured onboard semantics can support a conservative local
policy and trigger cloud tactical reasoning only when traffic conditions
warrant it.

\begin{figure}[!ht]
  \centering
  \includegraphics[width=\columnwidth]{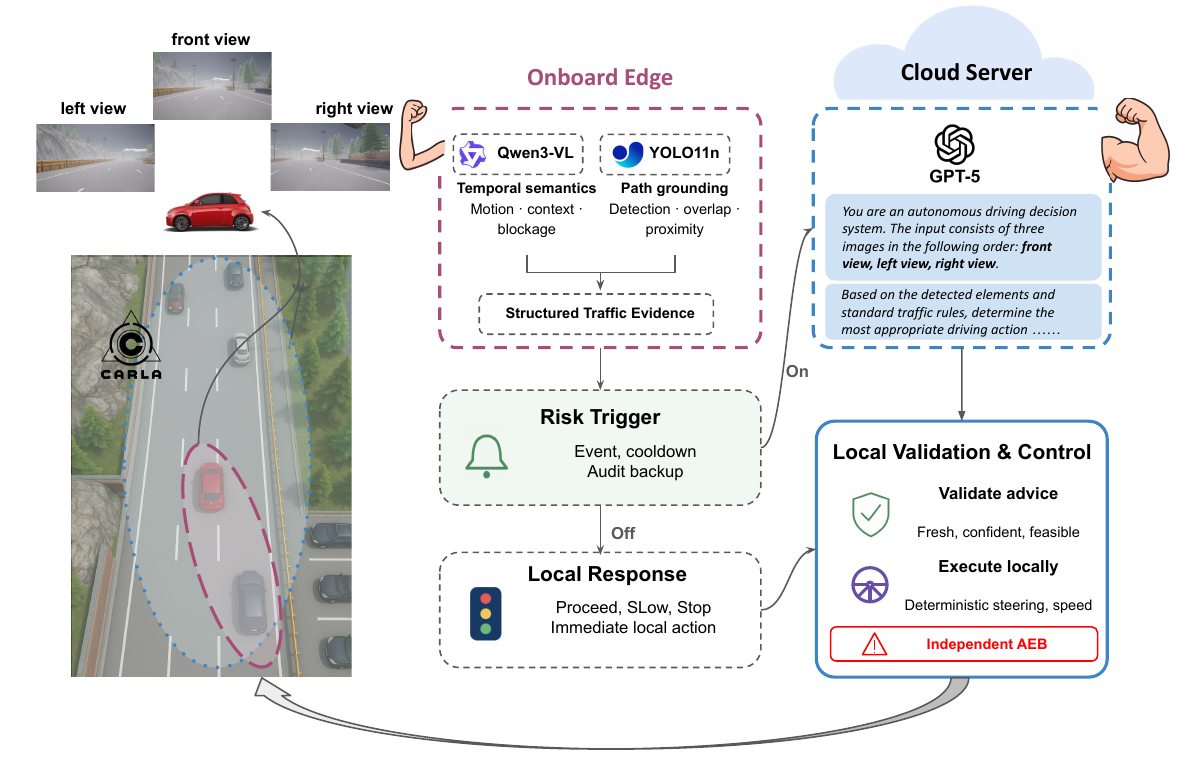}
  \caption{Asymmetric edge--cloud architecture. Onboard Qwen3-VL and YOLO11n
  produce temporal, path-grounded evidence for local policy and cloud
  scheduling. The cloud returns tactical advice, while validation, actuation,
  and emergency braking remain local.}
  \label{fig:system_overview}
\end{figure}

Figure~\ref{fig:system_overview} summarizes the proposed architecture. An
onboard VLM extracts temporal traffic semantics, and a lightweight detector
grounds supported road users relative to the planned lane. Deterministic logic
maps this evidence to a local speed policy and an urgency signal that governs
cloud requests. The cloud returns tactical advice rather than actuator
commands; freshness, confidence, and geometric feasibility are checked locally
before deterministic execution, and the shared AEB remains independent of the
learned models. Thus, the same structured evidence drives both local response
and selective cloud access.

We evaluate the system in a four-scenario policy comparison, a
delayed-roadwork event ablation, and three network profiles with added delay,
jitter, and loss. We report task completion, communication, AEB use, and
maneuver timing.

The paper makes three contributions:
\begin{itemize}

\item We frame edge--cloud driving as a joint scheduling-and-authority
problem. Structured onboard semantics govern both a conservative local policy
and cloud invocation, while advice validation, actuator control, and emergency
intervention remain local.

\item We implement a semantic escalation policy that combines temporal VLM
observations with planned-path grounding. It triggers requests when scene
urgency increases, rate-limits repeated requests under persistent hazards, and
retains a low-frequency audit to prevent indefinite cloud silence.

\item We measure the communication--responsiveness tradeoff in matched CARLA
experiments. In the main matrix, Risk-adaptive and
Periodic-cloud each completed 40 of 40 tasks, while Risk-adaptive issued
54.1\% fewer cloud requests; across the evaluated network profiles, lower
communication demand was accompanied by slower lane-change completion.

\end{itemize}

\section{Related Work}

Vision--language models have been incorporated into driving systems as
interfaces for scene understanding, decision making, instruction following,
and planning. DriveGPT4, GPT-Driver, and Talk2Drive use language-conditioned
representations to support driving decisions, explanations, or interaction
\cite{xu2024drivegpt4,mao2023gpt,cui2024personalized}. LMDrive studies
instruction-guided driving \cite{shao2024lmdrive}, while DriveVLM and VLP
integrate language-based reasoning into hierarchical driving and planning
pipelines \cite{tian2024drivevlm,pan2024vlp}. DriveLM instead organizes driving
reasoning as graph-structured visual question answering
\cite{sima2024drivelm}. These studies establish the value of semantic reasoning
within the driving stack. Our focus is not another end-to-end driving model,
but the runtime placement of that reasoning: when remote inference should be
requested and how its output should enter a locally controlled system.

Edge and split computing provide several mechanisms for distributing inference
across devices and remote resources. Neurosurgeon partitions neural networks
between device and cloud \cite{kang2017neurosurgeon}, while BranchyNet and
subsequent split-inference research study confidence-dependent exits and
device--cloud partitioning
\cite{teerapittayanon2016branchynet,xu2023survey,matsubara2022split}. In
transportation systems, edge computing and V2X architectures distribute
sensing and computation across vehicles, roadside infrastructure, and cloud
servers \cite{satyanarayanan2017emergence,gong2023edge,huang2023v2x,zhu2023cloud}.
Cooperative perception extends the observable field through shared vehicle and
roadside data \cite{zimmer2024tumtraf}, whereas EC-Drive selectively uploads
drift-critical observations for remote language-model reasoning
\cite{chen2024edge}. These systems make remote computation selective through
model partitioning, early exit, cooperative sensing, or observation upload.

We use an asymmetric policy rather than splitting the layers of a single
model. The onboard VLM and lightweight detector produce temporal,
path-relative evidence for both conservative local response and event-driven
cloud requests. The cloud returns tactical advice, which is checked locally
for freshness, confidence, and geometric feasibility before deterministic
execution. Our focus is the coupling between request timing and local
authority in cloud-assisted driving.

\section{Methodology}

\subsection{System Overview and Authority Boundary}

The proposed system separates semantic inference, tactical recommendation, and
vehicle actuation. As illustrated in Fig.~\ref{fig:system_overview}, the
onboard perception path produces structured temporal evidence that serves two
functions: it supports a conservative local driving policy and determines when
cloud reasoning should be requested. When invoked, the cloud adviser returns a
tactical recommendation. Local deterministic
modules check the recommendation and execute any accepted maneuver, while a
shared automatic emergency braking (AEB) system provides the final intervention
layer.

This design keeps actuation authority outside both learned models.
The onboard VLM reports observable traffic semantics, and the cloud VLM
recommends a maneuver; neither model directly sets steering, throttle, or
braking. The following subsections describe the onboard evidence representation,
the cloud-request policy, and the local validation and control process.

\subsection{Structured Onboard Semantic Evidence}

Risk-adaptive cloud access uses an onboard state that captures temporal
traffic behavior in a form that deterministic logic can interpret. Let
\begin{equation}
C_t=\{I_{t-k\Delta t}^{\mathrm{front}}\}_{k=0}^{N-1}
\end{equation}
denote a sequence of $N$ chronological front-camera frames sampled over a
temporal window of duration $T$, where $\Delta t=T/(N-1)$ is the sampling
interval. The onboard VLM maps this clip to a structured observation
\begin{equation}
\mathbf{z}_t=f_{\mathrm{edge}}(C_t).
\end{equation}
The fields in $\mathbf{z}_t$ describe path clearance, hazard category,
screen-space lateral position, relative motion, proximity, path blockage,
model confidence, and concise visible evidence. The prompt restricts the model
to observable scene properties. Edge inference runs asynchronously with a
minimum submission interval of 1~s; a new clip is not submitted while the edge
worker is occupied.

Screen-space position alone is insufficient to determine whether an observed
road user intersects the planned path. The system therefore projects the
planned lane into the latest front frame as corridor $\Gamma_t$ and
obtains supported road-user bounding boxes $\mathcal{B}_t$ from YOLO11n. For
each matched detection, the bottom center of the bounding box is compared with
$\Gamma_t$ to determine its path relevance. The detector provides this narrow
geometric grounding, while the onboard VLM provides temporal motion,
construction context, and semantic evidence for objects outside the detector's
supported classes.

A fixed mapping converts the combined evidence into the local semantic policy
\begin{equation}
r_t=\pi_{\mathrm{edge}}(\mathbf{z}_t,\mathcal{B}_t,\Gamma_t)
    \in\{\texttt{proceed},\texttt{slow},\texttt{stop}\}.
\end{equation}
An in-path hazard classified as approaching, crossing, near, or blocked maps to
\texttt{stop}, whereas a developing or uncertain conflict maps to
\texttt{slow}. After a visual stop,
the policy requires three consecutive clear observations before releasing the
stored stop state. This semantic latch is separate from the kinematic AEB
logic described below.

\subsection{Semantic Escalation and Cloud Tactical Advice}

The semantic scheduler converts the onboard traffic state into a cloud-request
decision. The local assessment is assigned an urgency level
\begin{equation}
U_t\in\{0,1,2,3\},
\end{equation}
corresponding respectively to no hazard, caution, stop, and blocked lane. Let
$d_t$ denote the elapsed time since the most recent cloud request. The
Risk-adaptive policy sets the request indicator $q_t$ as
\begin{equation}
q_t=
\begin{cases}
1, & \text{at startup},\\
1, & U_t>U_{t-1},\\
1, & U_t>0 \text{ and } d_t\geq3~\mathrm{s},\\
1, & d_t\geq10~\mathrm{s},\\
0, & \text{otherwise}.
\end{cases}
\end{equation}
An increase in urgency therefore triggers an immediate semantic-event request.
A persistent hazard can generate another request after the 3~s cooldown, while
the 10~s audit prevents indefinite cloud silence in the absence of a new
event. For comparison, the Periodic-cloud policy submits a request at startup
and approximately every 3~s without using semantic-event timing.

\begin{figure*}[!t]
  \centering
  \includegraphics[width=\textwidth]{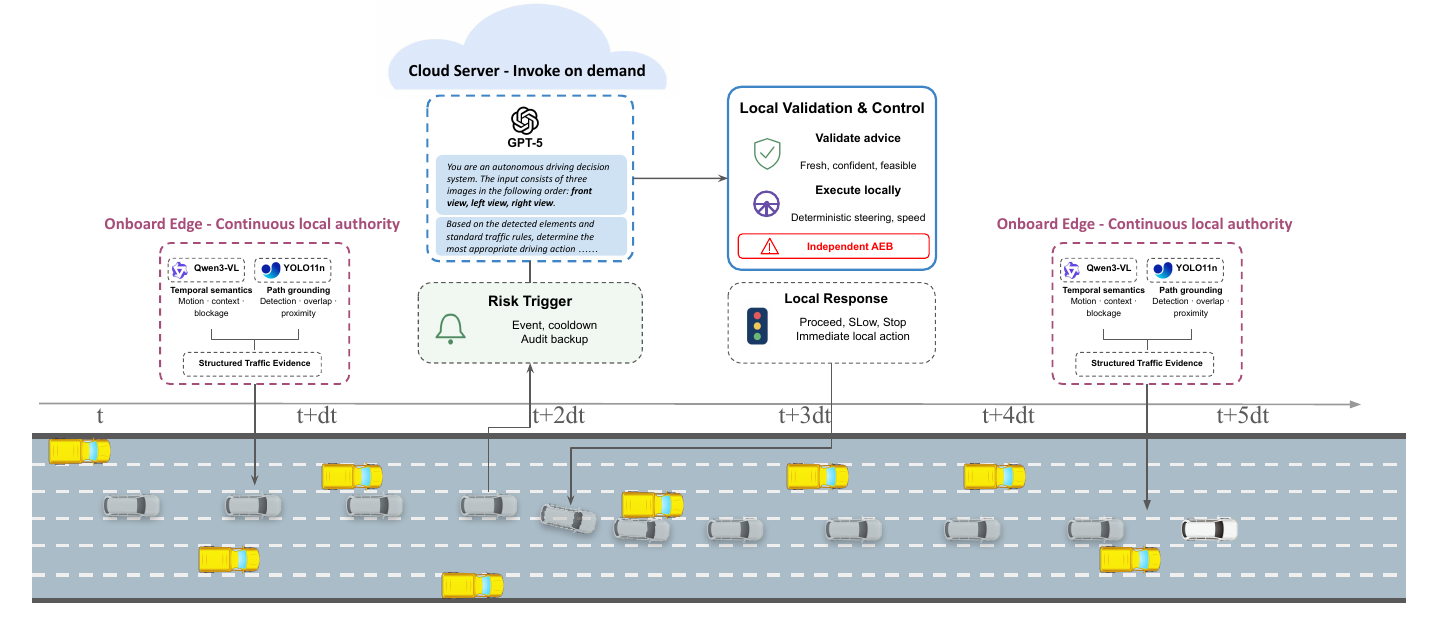}
  \caption{Temporal organization of onboard semantic inference and selective
  cloud reasoning. The onboard path updates structured traffic evidence and
  the local semantic policy. A cloud request is issued when the scheduler
  detects an increase in urgency, permits another request for a persistent
  hazard, or reaches the audit interval. Returned tactical advice is validated
  and executed locally. The displayed time spacing is schematic.}
  \label{fig:temporal_collaboration}
\end{figure*}

Figure~\ref{fig:temporal_collaboration} shows the resulting temporal
organization. Onboard semantic inference is repeated independently of cloud
access, whereas cloud reasoning occurs only when an event, hazard cooldown, or
audit condition is satisfied. The illustrated time points are schematic and do
not represent a mandatory cloud-query period.

When $q_t=1$, the cloud request contains left-, front-, and right-view JPEG
images, ego speed and acceleration, fresh onboard semantic evidence when the
selected policy includes the edge path, and the lane changes currently
admissible under local geometry. The cloud adviser returns
\begin{equation}
\mathbf{a}_t=(m_t,v_t^*,p_t,\ell_t,e_t),
\end{equation}
where $m_t$ denotes the recommended tactical maneuver, $v_t^*$ is an optional
target speed, $p_t$ is the reported confidence, $\ell_t$ is the proposed
validity duration, and $e_t$ is a concise explanation based on visible
evidence.

\subsection{Local Validation, Execution, and Shared AEB}

Cloud advice can affect the local driving plan only after passing confidence,
freshness, and feasibility checks. Recommendations with confidence below 0.60
are discarded. Non-lane-change advice is accepted for at most 2~s after its
generation, whereas lane-change advice is accepted for at most 5~s. In both
cases, the effective lifetime is the smaller of this local limit and the
cloud-provided duration $\ell_t$. A cloud request times out after 5~s and is
not automatically retried.

Lane-change recommendations receive an additional geometric check immediately
before execution. The target lane must exist, have the same travel direction,
and provide at least 20~m of front clearance and 10~m of rear clearance. Once
initiated, the maneuver remains active beyond the lifetime of the originating
cloud response, uses a target speed no greater than 4~m/s, and terminates after
at most 8~s. A fresh dynamic edge stop can veto the maneuver. In contrast, a
static lane-closure observation can support continuation of a locally validated
bypass.

Accepted tactical decisions are executed by deterministic controllers. A
pure-pursuit controller tracks the selected lane waypoint, and a proportional
controller tracks the target speed. All evaluated policies additionally use
the same AEB system. Let $g_t$ denote the surface gap to the relevant obstacle
and $\mathrm{TTC}_t$ its time to collision. The AEB replaces the nominal
command with full braking when
\begin{equation}
g_t\leq5~\mathrm{m}
\quad\text{or}\quad
0\leq\mathrm{TTC}_t\leq1.5~\mathrm{s}.
\end{equation}

For outcome attribution, an event is assigned to the upstream edge or cloud
decision when that decision has already commanded braking equivalent to the
AEB response; under this convention, the event is not recorded as an AEB
activation. The reported strict-success metric is computed using this recorded
activation variable. Because $g_t$ and $\mathrm{TTC}_t$ are obtained from
CARLA ground-truth actor states, the AEB is an experimental backstop rather
than a deployable perception-based safety system.

\section{Experimental Evaluation}

\subsection{Setup and Protocol}

We evaluate all policies in CARLA~0.9.15~\cite{dosovitskiy2017carla}, using
Town04 and a Tesla Model~3. Each 30~s run uses a 10~Hz simulator and control
loop, an 11~m/s cruise speed, and a $640\times360$ camera with a $90\degree$
field of view. The edge stack comprises Qwen3-VL-4B-Instruct
\cite{bai2025qwen3vl} and CPU-based YOLO11n~\cite{jocher2024yolo11}; the cloud
adviser uses \texttt{gpt-5-mini} with minimal reasoning and a 5~s timeout.
Edge and cloud inference are asynchronous and do not block vehicle control.

The implementation retains detector, semantic, and tactical-advice confidence
as separate quantities. YOLO11n detections use the detector-native confidence
score and are filtered at 0.25. The edge VLM reports an uncalibrated semantic
self-assessment score $c_t^{\mathrm{sem}}\in[0,1]$ for the complete structured
observation; scores below 0.50 follow the conservative slow path. The cloud
adviser similarly reports an uncalibrated recommendation score
$p_t\in[0,1]$, and advice below 0.60 is discarded. Advice above this threshold
must still satisfy freshness and local geometric-feasibility checks. Detector
and model-reported scores are logged separately and are not combined through a
weighted confidence-fusion rule.

Table~\ref{tab:modes} summarizes the four policies. Edge-only performs semantic
perception and conservative response locally. Periodic-cloud omits the edge
VLM and queries the cloud at startup and approximately every 3~s.
Risk-adaptive combines the edge stack with startup, semantic-event, and 10~s
audit requests. Audit-only, used only in RQ2, removes semantic-event requests
while retaining the Risk-adaptive edge stack and audit schedule. All policies
share the deterministic controller, lane validation, and AEB.

\begin{table}[t]
\centering
\caption{Compared access policies; Audit-only is used only in RQ2.}
\label{tab:modes}
\scriptsize
\setlength{\tabcolsep}{2.5pt}
\begin{tabularx}{\columnwidth}{@{}l
>{\raggedright\arraybackslash}X
>{\raggedright\arraybackslash}X@{}}
\toprule
Policy & Edge semantics & Request schedule \\
\midrule
Edge-only & Qwen3-VL, YOLO, local mapping & None \\
Periodic-cloud & None & Startup and $\sim$3~s periodic \\
Audit-only & Same as Risk-adaptive & Startup and 10~s audit \\
Risk-adaptive & Qwen3-VL, YOLO, local mapping & Startup, 10~s audit, semantic events \\
\bottomrule
\end{tabularx}
\end{table}

Task success requires completion of the scenario objective without collision:
collision-free traversal for clear-road and adjacent-lane trials, a final
speed $\leq0.5$~m/s for static obstacles, and passage of the closure for
roadwork. Strict success further excludes any recorded AEB activation.
Secondary measures include minimum surface gap and TTC where defined,
roadwork maneuver timing, cloud requests, JPEG payload, token use, and response
latency. In nonblocking scenes, unnecessary intervention is measured as time
after 3~s with a commanded target speed $\leq0.5$~m/s.

\section{Results}

\subsection{RQ1: Overall Effectiveness and Communication}

RQ1 compares Edge-only, Periodic-cloud, and Risk-adaptive in four scenarios:
clear road, an adjacent-lane vehicle, a static in-lane obstacle, and a roadwork
merge. Ten matched seeds and obstacle distances yield 40 runs per policy and
120 valid runs. Each scenario targets a different behavior. Clear-road trials
measure unnecessary intervention; adjacent-lane trials test whether path
grounding rejects a nearby but nonconflicting vehicle; static-obstacle trials
test conservative longitudinal stopping; and roadwork trials require a
tactical lane change rather than longitudinal slowing alone. Seeds and
obstacle distances are matched across policies within each scenario, while
the controller, lane validator, and AEB remain unchanged. Both cloud-capable
policies completed all 40 tasks
(Table~\ref{tab:exp1_outcomes}); Edge-only completed 30/40, failing all ten
roadwork bypasses. Risk-adaptive recorded 35 strict successes and five AEB
activations, compared with 29 and 11 for Periodic-cloud. Because the latter
omits the edge VLM, these differences characterize the complete policy stacks
rather than the causal effect of request scheduling alone.

At the scenario level, Edge-only completed the
clear-road, adjacent-lane, and static-obstacle trials but stopped before every
roadwork closure. Its local response could slow or stop the vehicle in these
trials, but did not supply the tactical lane change needed to pass the closure.

\begin{table}[t]
\centering
\caption{RQ1 outcomes over 40 runs per policy. Strict success excludes recorded
AEB activation.}
\label{tab:exp1_outcomes}
\scriptsize
\setlength{\tabcolsep}{2pt}
\begin{tabular}{@{}lrrrr@{}}
\toprule
Policy & Task $\uparrow$ & Strict $\uparrow$ & AEB $\downarrow$ & Crash $\downarrow$ \\
\midrule
Edge-only & 30/40 (75.0\%) & 24/40 (60.0\%) & 11/40 & 0/40 \\
Periodic-cloud & 40/40 (100\%) & 29/40 (72.5\%) & 11/40 & 0/40 \\
Risk-adaptive & 40/40 (100\%) & 35/40 (87.5\%) & 5/40 & 0/40 \\
\bottomrule
\end{tabular}
\end{table}

\begin{figure}[!t]
  \centering
  \includegraphics[width=\columnwidth]
    {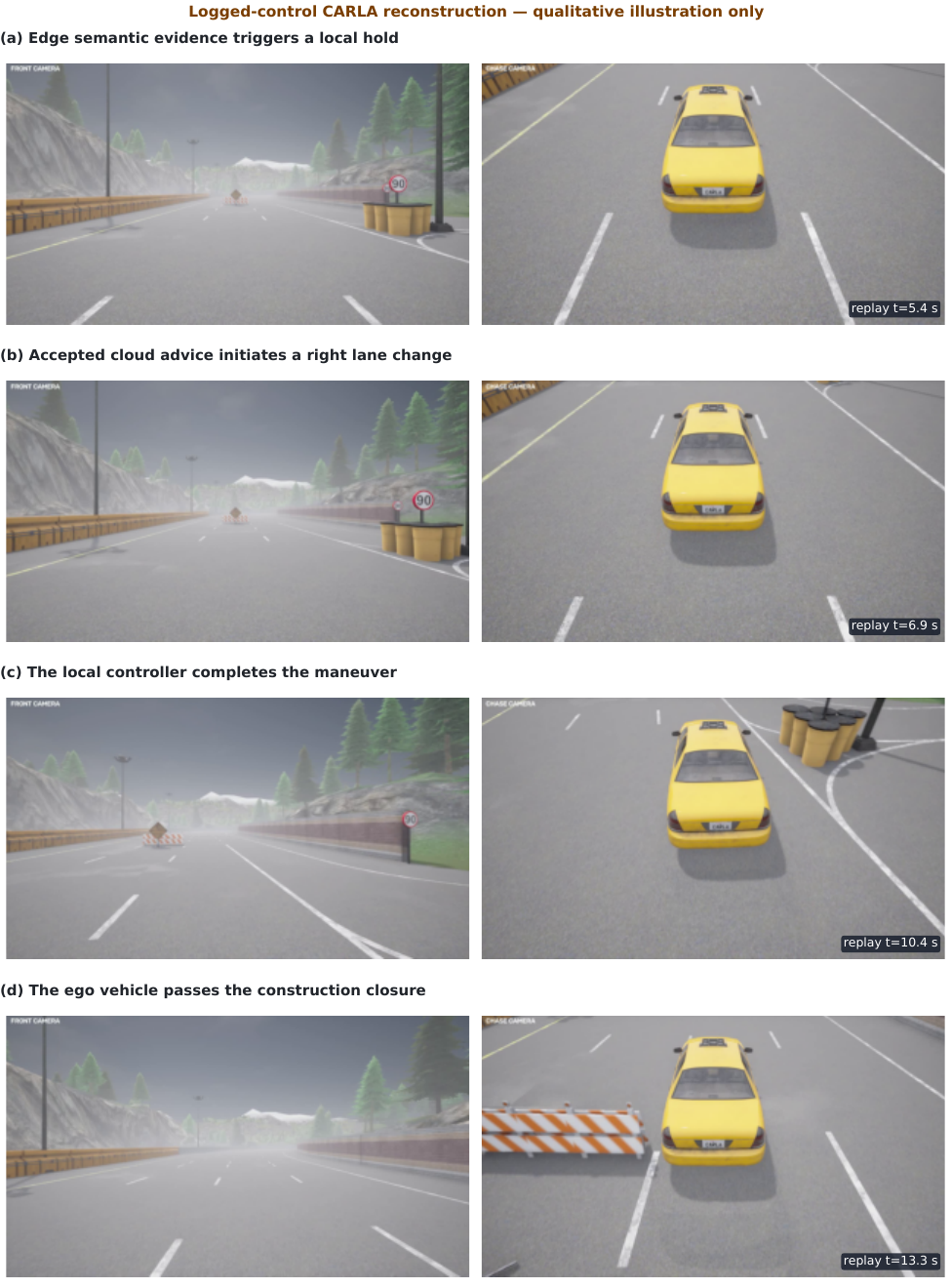}
  \caption{Qualitative reconstruction of a Risk-adaptive roadwork maneuver:
  (a) local semantic hold, (b) accepted lane-change advice, (c) local maneuver
  execution, and (d) passage of the closure. Images are reconstructed from a
  logged control trace rather than original experiment footage.}
  \label{fig:roadwork_sequence}
\end{figure}

Figure~\ref{fig:roadwork_sequence} shows one successful roadwork run. Edge
semantics first induce a local hold, accepted cloud advice selects a bounded
lane change, and the local controller executes the maneuver and passes the
closure. The figure is a
reconstruction from the logged 10~Hz control trace and is not counted as an
additional experiment.

Risk-adaptive used 54.1\% fewer cloud requests, 51.9\% less image payload,
and 51.2\% fewer tokens than Periodic-cloud (Table~\ref{tab:exp1_comm}). Savings
were largest in nonblocking scenes; a persistent static hazard elicited more
event-driven requests. Request density followed semantic state rather than
approximating a lower fixed polling rate. 

The communication reduction arose primarily from fewer cloud invocations
rather than smaller individual requests. Mean image payload was
0.130~MB per Periodic-cloud request and 0.137~MB per Risk-adaptive request,
while total token use averaged 1,376 and 1,463 tokens per request,
respectively. Thus, semantic escalation changes when cloud reasoning is
invoked; it does not compress the multiview request or reduce its per-request
reasoning demand.

\begin{table}[t]
\centering
\caption{RQ1 aggregate cloud communication.}
\label{tab:exp1_comm}
\footnotesize
\setlength{\tabcolsep}{3pt}
\begin{tabular}{lrrr}
\toprule
Policy & Calls $\downarrow$ & Payload (MB) $\downarrow$ & Tokens $\downarrow$ \\
\midrule
Periodic-cloud & 399 & 52.011 & 549,011 \\
Risk-adaptive & 183 & 25.043 & 267,786 \\
Reduction & 54.1\% & 51.9\% & 51.2\% \\
\bottomrule
\end{tabular}
\end{table}

\subsection{RQ2: Contribution of Semantic-Event Requests}

RQ2 isolates event-triggered access in delayed roadwork, where the closure is
revealed after 5~s. Five matched closure distances yield 15 runs. Audit-only
and Risk-adaptive differ only in whether a semantic event can trigger a cloud
request; Periodic-cloud provides a denser-access reference. Risk-adaptive
completed 5/5 tasks, versus 4/5 for Audit-only and 3/5 for Periodic-cloud
(Table~\ref{tab:exp2}). It completed the lane change earlier than Audit-only in
four matched cases, with a mean paired advantage of 3.86~s and a median of
1.30~s; Audit-only was 0.8~s faster in the remaining case.
Risk-adaptive required seven more calls than Audit-only but 56.0\% fewer than
Periodic-cloud. These five cases support the intended timing mechanism but are
insufficient to establish a general performance or safety advantage. The
longest Audit-only response missed completion within the 30~s horizon, while
one matched Audit-only run completed sooner than Risk-adaptive.

\begin{table}[t]
\centering
\caption{RQ2 delayed-roadwork ablation (five runs per policy).}
\label{tab:exp2}
\scriptsize
\setlength{\tabcolsep}{2.2pt}
\begin{tabular}{@{}lrrrrr@{}}
\toprule
Policy & Task $\uparrow$ & Strict $\uparrow$ & AEB $\downarrow$ &
Calls $\downarrow$ & Payload (MB) $\downarrow$ \\
\midrule
Periodic-cloud & 3/5 & 3/5 & 2 & 50 & 6.652 \\
Audit-only & 4/5 & 3/5 & 1 & 15 & 2.018 \\
Risk-adaptive & 5/5 & 4/5 & 1 & 22 & 2.903 \\
\bottomrule
\end{tabular}
\end{table}

\begin{figure}[t]
  \centering
  \includegraphics[width=\columnwidth]
    {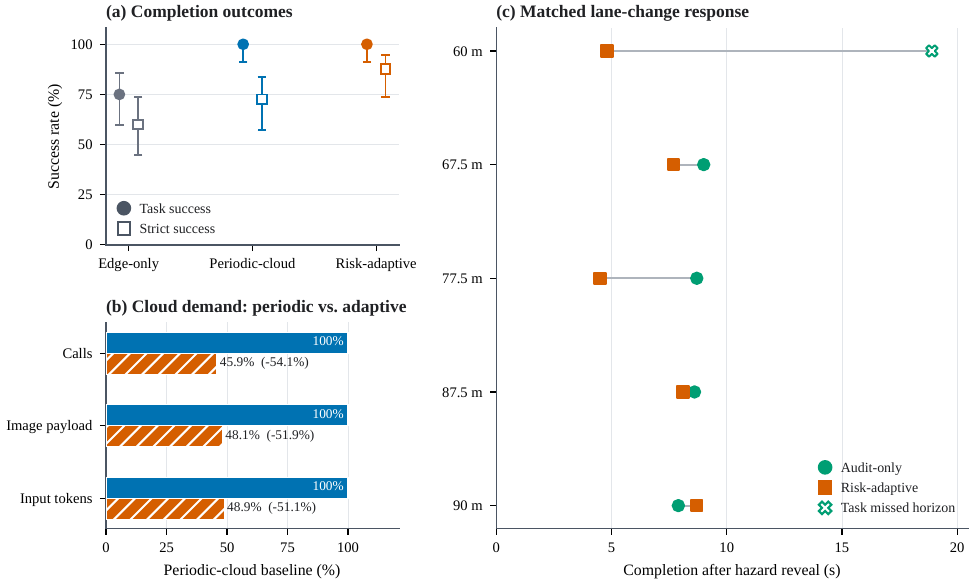}
  \caption{RQ1--RQ2 evidence. (a) Task and strict success with Wilson 95\%
  confidence intervals. (b) Cloud demand normalized to Periodic-cloud; Edge-only
  is omitted because it makes no cloud requests. (c) Matched reveal-to-lane-change
  times in RQ2; the cross denotes a task not completed within the horizon.}
  \label{fig:rq1_rq2_evidence}
\end{figure}

\subsection{RQ3: Communication--Responsiveness Tradeoff}

RQ3 compares Periodic-cloud and Risk-adaptive on delayed roadwork under Normal,
Moderate, and Severe network profiles. Moderate adds 500~ms latency and 100~ms
Gaussian jitter; Severe adds 1000~ms latency, 200~ms jitter, and 10\% emulated
loss. Each policy--profile cell contains 20 trials balanced across five closure
distances (120 entries). Because model outputs and API latency remain
stochastic, the profiles do not isolate network effects through deterministic
replay.

Risk-adaptive made fewer calls under every evaluated profile. Relative to
Periodic-cloud, it reduced calls by 47.2\%, 48.0\%, and
41.1\% under Normal, Moderate, and Severe conditions, respectively; the
corresponding payload reductions were 45.3\%, 46.4\%, and 39.7\%. The smaller
reduction under Severe conditions occurred alongside the largest
lane-completion delay.

Both policies completed 47/60 tasks in aggregate (Table~\ref{tab:exp3}).
Risk-adaptive reduced calls from 587 to 320 (45.5\%) and payload from 75.797 to
42.630~MB (43.8\%), but its mean lane-change completion time was 1.92--3.67~s
longer across profiles. It also recorded 38/60 strict successes and 12 AEB
activations, compared with 32/60 and 28 for Periodic-cloud. The lower
communication demand therefore came with slower tactical response; the
profile-level task-success differences remain inconclusive.

\begin{table}[!t]
\centering
\caption{RQ3 delayed-roadwork results. Success is count/20 [Wilson 95\% CI];
lane time is the mean among completed maneuvers.}
\label{tab:exp3}
\scriptsize
\setlength{\tabcolsep}{2.6pt}
\renewcommand{\arraystretch}{1.08}
\textit{(a) Outcomes}\par\vspace{2pt}
\begin{tabular}{@{}llccc@{}}
\toprule
Profile & Policy & Task $\uparrow$ & Strict $\uparrow$ & AEB $\downarrow$ \\
\midrule
Normal & Periodic & \shortstack{18/20\\{\tiny [70--97]}} &
\shortstack{17/20\\{\tiny [64--95]}} & 3 \\
& Risk-adaptive & \shortstack{17/20\\{\tiny [64--95]}} &
\shortstack{16/20\\{\tiny [58--92]}} & 1 \\
\addlinespace[2pt]
Moderate & Periodic & \shortstack{16/20\\{\tiny [58--92]}} &
\shortstack{12/20\\{\tiny [39--78]}} & 8 \\
& Risk-adaptive & \shortstack{17/20\\{\tiny [64--95]}} &
\shortstack{12/20\\{\tiny [39--78]}} & 6 \\
\addlinespace[2pt]
Severe & Periodic & \shortstack{13/20\\{\tiny [43--82]}} &
\shortstack{3/20\\{\tiny [5--36]}} & 17 \\
& Risk-adaptive & \shortstack{13/20\\{\tiny [43--82]}} &
\shortstack{10/20\\{\tiny [30--70]}} & 5 \\
\bottomrule
\end{tabular}

\par\vspace{5pt}
\textit{(b) Communication and maneuver time}\par\vspace{2pt}
\begin{tabular}{@{}llrrr@{}}
\toprule
Profile & Policy & Calls $\downarrow$ & MB $\downarrow$ & Lane (s) $\downarrow$ \\
\midrule
Normal & Periodic & 199 & 25.4 & 5.5 \\
& Risk-adaptive & 105 & 13.9 & 7.4 \\
Moderate & Periodic & 196 & 25.2 & 6.0 \\
& Risk-adaptive & 102 & 13.5 & 8.8 \\
Severe & Periodic & 192 & 25.2 & 7.5 \\
& Risk-adaptive & 113 & 15.2 & 11.2 \\
\bottomrule
\end{tabular}
\end{table}

The failure patterns identify different bottlenecks. All 13 Periodic-cloud
failures lacked a completed lane change and ended with AEB activation. In nine
of 13 Risk-adaptive failures, the vehicle completed or substantially executed
the evasive maneuver but did not pass the closure within 30~s, commonly because
conservative roadwork semantics remained latched. Thus, fewer recorded AEB
activations should not be interpreted as a safety result: post-maneuver
semantic release can still prevent task completion.

\section{Discussion and Limitations}

The edge and cloud components had different roles in the evaluated stack.
Temporal and path-relative edge semantics governed conservative local response
and request timing. The cloud selected among locally admissible tactical
maneuvers, subject to local validation and control. In RQ2, semantic events
advanced a request that would otherwise wait for the next audit. In RQ1 and
RQ3, selective escalation approximately halved cloud demand, while periodic
access produced faster lane changes when frequent advice was useful.

The lightweight detector and edge VLM also serve different functions. The
detector grounds supported road users relative to the planned path; the VLM
provides temporal motion and construction context to the local policy and
request scheduler. RQ2 examines the scheduling consequence of this semantic
state, not the performance of either component against a detector-only stack.

Task success, strict success, and AEB activation describe different outcomes.
The shared AEB can rescue late upstream decisions, while a conservative stop
after an evasive maneuver can avoid AEB yet still fail the task. The AEB counts
diagnose upstream behavior under this simulator protocol; they do
not establish real-world safety. The RQ3 failures further suggest that
improving path-relative semantic clearing may be as important as increasing
request frequency.

The study is limited to one CARLA town and ego vehicle, scripted scenarios,
and 30~s horizons. Half of RQ1 informed development, RQ2 contains five matched
cases, and RQ3 reuses ten entries. Stochastic model and API behavior prevents
causal isolation of the injected network conditions, the confidence intervals
are broad, and no formal noninferiority claim is made. Moreover, lane
projection, maneuver feasibility, and AEB use simulator geometry, while the
reported resource measures omit edge compute, energy, monetary cost, and
certified latency. The scenarios also exclude pedestrians, signals, dense
traffic, adverse weather, localization noise, and unstructured roads.
Evaluation with sensor-derived geometry, deterministic network replay, broader
traffic conditions, and real communication links is needed before drawing
deployment or safety conclusions.

\section{Conclusion}

Risk-adaptive and periodic access both completed 40/40 tasks in the main
matrix, while Risk-adaptive used 54.1\% fewer cloud requests. The
delayed-roadwork ablation showed that semantic events could advance a request,
and the network study showed that lower cloud demand came with slower lane
changes. Under the tested simulator conditions, structured edge semantics
concentrated cloud access around relevant events without reducing aggregate
task completion. Future experiments should address semantic release after a
maneuver and use deterministic replay to isolate network effects before
evaluation on sensor-based driving stacks.

\clearpage

\bibliographystyle{IEEEtran}
\bibliography{trb2027_references_checked}

\end{document}